\documentclass[10pt,twocolumn,letterpaper]{article}

\usepackage{cvpr}
\usepackage{times}
\usepackage{epsfig}
\usepackage{graphicx}
\usepackage{amsmath}
\usepackage{amssymb}

\usepackage{adjustbox}
\newcommand{\widthscalefive}{0.120}
\newcommand{\widthscalefour}{0.160}
\usepackage{animate}

\usepackage{colortbl}
\usepackage{booktabs}
\usepackage[table,x11names,dvipsnames,xcdraw]{xcolor}

\usepackage[pagebackref=true,breaklinks=true,letterpaper=true,colorlinks,bookmarks=false]{hyperref}

\cvprfinalcopy % *** Uncomment this line for the final submission

\def\cvprPaperID{5932} % *** Enter the CVPR Paper ID here

\ifcvprfinal\fi
\begin{document}

%%%%%%%%% TITLE
\title{Unfolding a blurred image}

\author{Kuldeep Purohit$^{1}$ \qquad Anshul Shah$^{2}\thanks{Work done while at Indian Institute of Technology Madras, India.}$ \qquad A. N. Rajagopalan$^{1}$ \\
$^1$ Indian Institute of Technology Madras, India \hspace{1cm}
$^2$ University of Maryland, College Park\\
%{\tt\small \{chi.zhang,f.gao,baoxiongjia,yixin.zhu\}@ucla.edu, sczhu@stat.ucla.edu}
{\tt\small kuldeeppurohit3@gmail.com, anshulb@cs.umd.edu, raju@ee.iitm.ac.in}
} 

%%\author{Kuldeep Purohit, Anshul Shah, and A N Rajagopalan\\
%%Indian Institute of Technology Madras, India\\
%%%Institution1 address\\
%%{\tt\small  \{ee14s007,ee13b068,raju\}@ee.iitm.ac.in}}

%\and

%\author{Kuldeep Purohit, Anshul Shah, A N Rajagopalan}
%\institute{Indian Institute of Technology Madras, India\\
%	\email{ \{ee14s007,ee13b068,raju\}@ee.iitm.ac.in}
%}

\maketitle

\begin{abstract}
We present a solution for the goal of extracting a video from a single motion blurred image to sequentially reconstruct the clear views of a scene as beheld by the camera during the time of exposure. We first learn motion representation from sharp videos in an unsupervised manner through training of a convolutional recurrent video autoencoder network that performs a surrogate task of video reconstruction. Once trained, it is employed for guided training of a motion encoder for blurred images. This network extracts embedded motion information from the blurred image to generate a sharp video in conjunction with the trained recurrent video decoder. As an intermediate step, we also design an efficient architecture that enables real-time single image deblurring and outperforms competing methods across all factors: accuracy, speed, and compactness. Experiments on real scenes and standard datasets demonstrate the superiority of our framework over the state-of-the-art and its ability to generate a plausible sequence of temporally consistent sharp frames. 
\end{abstract}

\section{Introduction}

Recent works on future frame prediction reveal that direct intensity estimation leads to blurred predictions. Instead, if a frame is reconstructed based on the original image and corresponding transformations, both scene dynamics and invariant appearance can be preserved well. Based on this premise, \cite{flynn2016deepstereo,zhou2016view} and \cite{liu2017video} model the task as a flow of image pixels. The methods \cite{vondrick2016generating,xue2016visual} generate a video from a single sharp image, but have a severe limitation in that they work only on the specific scene for which they are trained. All of these approaches work only on sharp images and videos. However, motion during exposure is known to cause severe degradation in the captured image quality due to the blur it induces. This is usually the case in low-light situations where the exposure time of each frame is high and in scenes where significant motion happens within the exposure time. In \cite{vasiljevic2016examining}, it has been shown that standard network models used for vision tasks and trained only on high-quality images suffer a significant degradation in performance when applied to images degraded by blur.

Motion deblurring is a challenging problem in computer vision due to its ill-posed nature. Recent years have witnessed significant advances in deblurring \cite{vasu2017local,pan2016blind,pan2014deblurring,chandramouli2010inferring,paramanand2011depth,paramanand2013non,vijay2013non,paramanand2014shape,rao2014inferring,rao2014harnessing,nimisha2017blur,vasu2017local,nimisha2018unsupervised,nimisha2018generating,vasu2018non,purohit2018learning,purohit2019bringing,mohan2019unconstrained,mohan2019unconstrained,purohit2020region,mohan2021deep}. Several methods \cite{xu2010two,pan2016robust,fergus2006removing,shan2008high,cho2009fast,joshi2008psf,krishnan2009fast,krishnan2011blind,xu2010two} have been proposed to address this problem using hand-designed priors as well as Convolutional Neural Networks (CNN)  \cite{chakrabarti2016neural,schuler2013machine,schuler2016learning}  for recovering the latent image. A few methods \cite{sun2015learning,gong2017motion} have been proposed to remove heterogeneous blur but they are limited in their capability to handle general dynamic scenes. Most of these methods strongly rely on the accuracy of the assumed image degradation model and include intensive, sometimes heuristic, parameter-tuning and expensive computations, factors which severely restrict their accuracy and applicability in real-world scenarios. The recent works of \cite{nah2017deep,nimisha2017blur,kupyn2017deblurgan,tao2018scale} overcome these limitations to some extent by learning to directly generate the latent sharp image, without the need for blur kernel estimation. 

We present a two-stage deep convolutional architecture to carve out a video from a motion blurred image that is applicable to non-uniform motion caused by individual or combined effects of camera motion, object motion and arbitrary depth variations in the scene. We avoid overly simplified models to represent motion and hence refrain from creating synthetic datasets for supervised training. The first stage consists of training a video auto-encoder wherein the encoder accepts a sequence of video frames to extract a latent motion representation while the decoder estimates the same video by applying estimated motion trajectories to a single sharp frame in a recurrent fashion. We use this trained video decoder to guide the training of a CNN (which we refer to as Blurred Image Encoder (BIE)) to extract the same motion information from a blurred image as the video encoder would from the image sequence corresponding to that blurred image. For testing, we propose an efficient deblurring network to first estimate a sharp frame from the given blurred image. The BIE is responsible for extracting motion features from the blurred image. The video decoder uses the outputs of the BIE and the deblurred sharp frame to generate the video underlying the motion blurred image.

As the only other work of this kind, \cite{jin2018learning} very recently proposed a method to estimate a video from a single blurred image by training multiple neural networks to estimate the underlying frames. In contrast, our architecture utilizes a single recurrent neural network to generate the entire sequence. Our recurrent design implicitly addresses temporal ambiguity to a large extent, since generation of any frame in the sequence is naturally preconditioned on all the previous frames. The approach of \cite{jin2018learning} is limited to small motion, owing to its architecture and training procedure. We estimate pixel level motion instead of intensities which proves to be an advantage for the task at hand, especially in cases with large blur (which is an issue with \cite{jin2018learning}). Our deblurirng architecture not only outperforms all existing deblurring methods but is also smaller and significantly faster. In fact, separating the processes of content and motion estimation allows our architecture to be used with any off-the-shelf deblurring approach. 

\section{The Proposed Architecture}
\label{sec:proposed architecture}

\begin{figure*}
\begin{center}
   \includegraphics[width=0.8\linewidth]{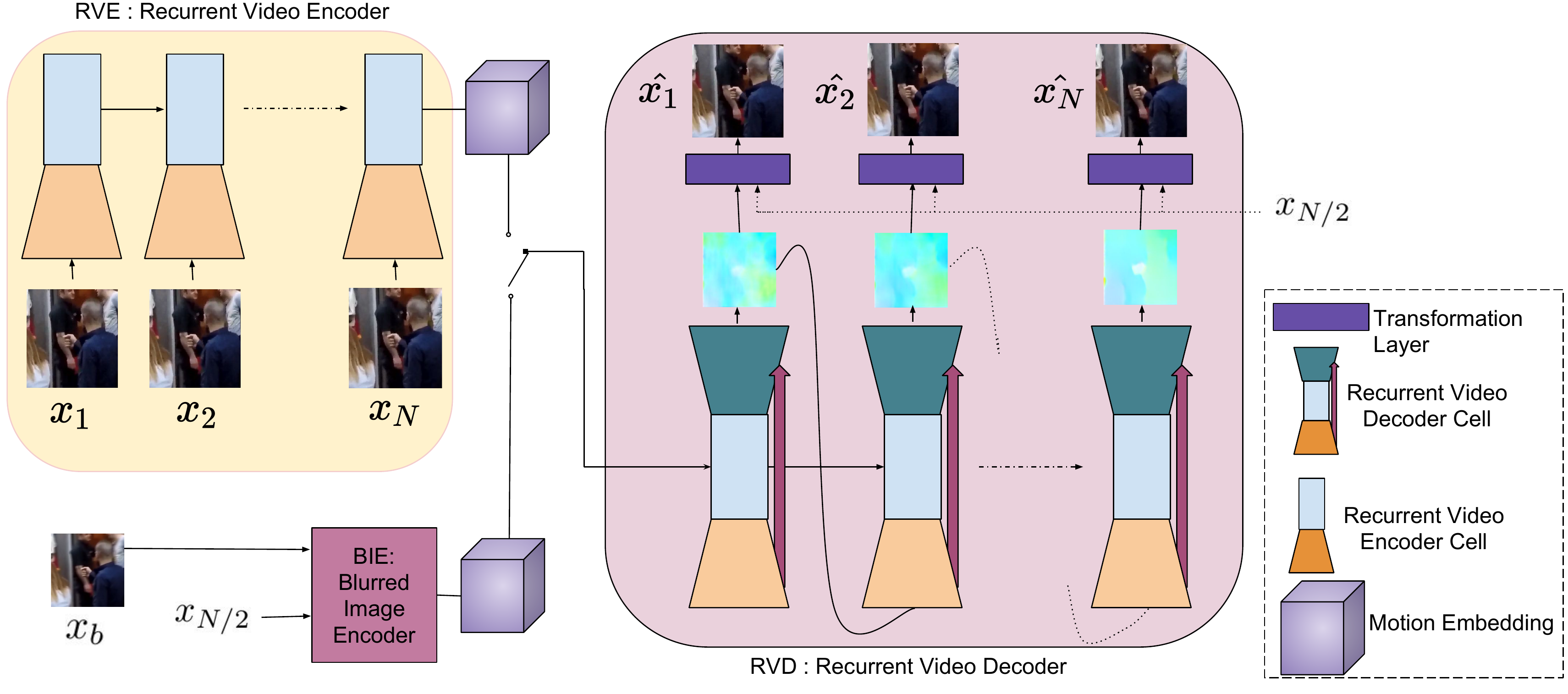}
\end{center}
   \caption{An overview of our video generation architecture during training. The first step involves training the RVE-RVD for the task of video reconstruction. This is followed by guided training of BIE through the trained RVD.}
\label{fig:architecture}
\end{figure*}

In our proposed video autoencoder, the encoder utilizes all the video frames to extract a latent representation, which is then fed to decoder which estimates the frame sequence in a recurrent fashion. The Recurrent Video Encoder (RVE) reads $N$ sharp frames ${x}_{1..N}$, one at each time-step. It returns a tensor at the last time-step, which is utilized as the motion representation of the image sequence. This tensor is used to initialize the first hidden state of another ConvLSTM based network called Recurrent Video Decoder (RVD) whose task is to recurrently estimate $N$ optical flows. Since the RVE-RVD pair is trained using reconstruction loss between the estimated frames $\hat{x}_{1..N}$ and ground-truth frames ${x}_{1..N}$, the RVD must return the predicted video. To enable this, the (known) central frame of the video is acted upon by the flows predicted by the RVD. Specifically, the estimated flows are individually fed to a differentiable transformation layer to transform the central frame $x_{\lfloor\frac{N}{2}\rfloor}$ to obtain the frames $\hat{x}_{1..N}$. Once trained, we have an RVD which can estimate sequential motion flows, given a particular motion representation.

In addition, we introduce another network called Blurred Image Encoder (BIE) whose task is to accept blurred image $x_B$ corresponding to the spatio-temporal average of the input frames ${x}_{1..N}$ and return a motion encoding, which too can be used to generate a sharp video.  
To achieve this task, we employ the already trained RVD to guide the training of BIE so as to extract the same motion information from the blurred image as the RVE would from that image sequence. In other words, the weights are to be learnt such that $BIE(x_B) \approx RVE(x_{1..N})$. We refrain from using the encoding returned by RVE for training due to lack of ground truth for the encoded representation. Instead, the BIE is trained such that the predicted video at the output of RVD for the given $x_B$ matches as closely as possible to the ground truth frames ${x}_{1..N}$. This ensures that the BIE learns to capture ordered motion information for the RVD to return a realistic video. Directly training the BIE-RVD pair poses a challenge since it requires learning to perform two tasks jointly: ``video generation from motion representation'' and ``ambiguity-invariant motion extraction from a blurred image''. Such training delivers below-par performance (see supplementary material).% Directly training the BIE-RVD pair is cumbersome and may lead to the RVD learning to extract motion whereas this task is to be performed by RVE/BIE.

The overall architecture of the proposed methodology is given in Fig. \ref{fig:architecture}. It is fully convolutional, end-to-end differentiable and can be trained using unlabeled high frame-rate videos, without the need for optical flow supervision, which is challenging to produce at large scale. During \emph{testing}, the central sharp frame is not available and is estimated using an independently trained deblurring module (DM). We now describe the design aspects of the different modules.

\subsection{Recurrent Video Encoder (RVE)}
\label{sec:rve}

\begin{figure}
\begin{center}
 \begin{tabular}{cc}
   \includegraphics[width=0.24\textwidth]{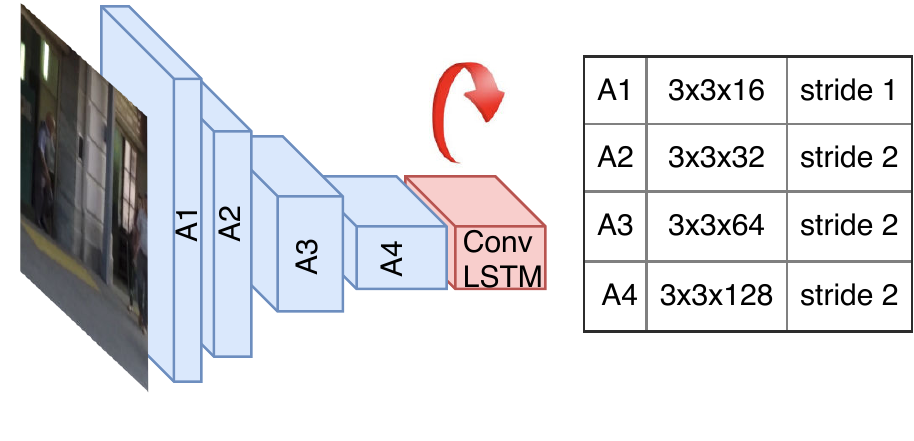} &
      \includegraphics[width=0.20\textwidth]{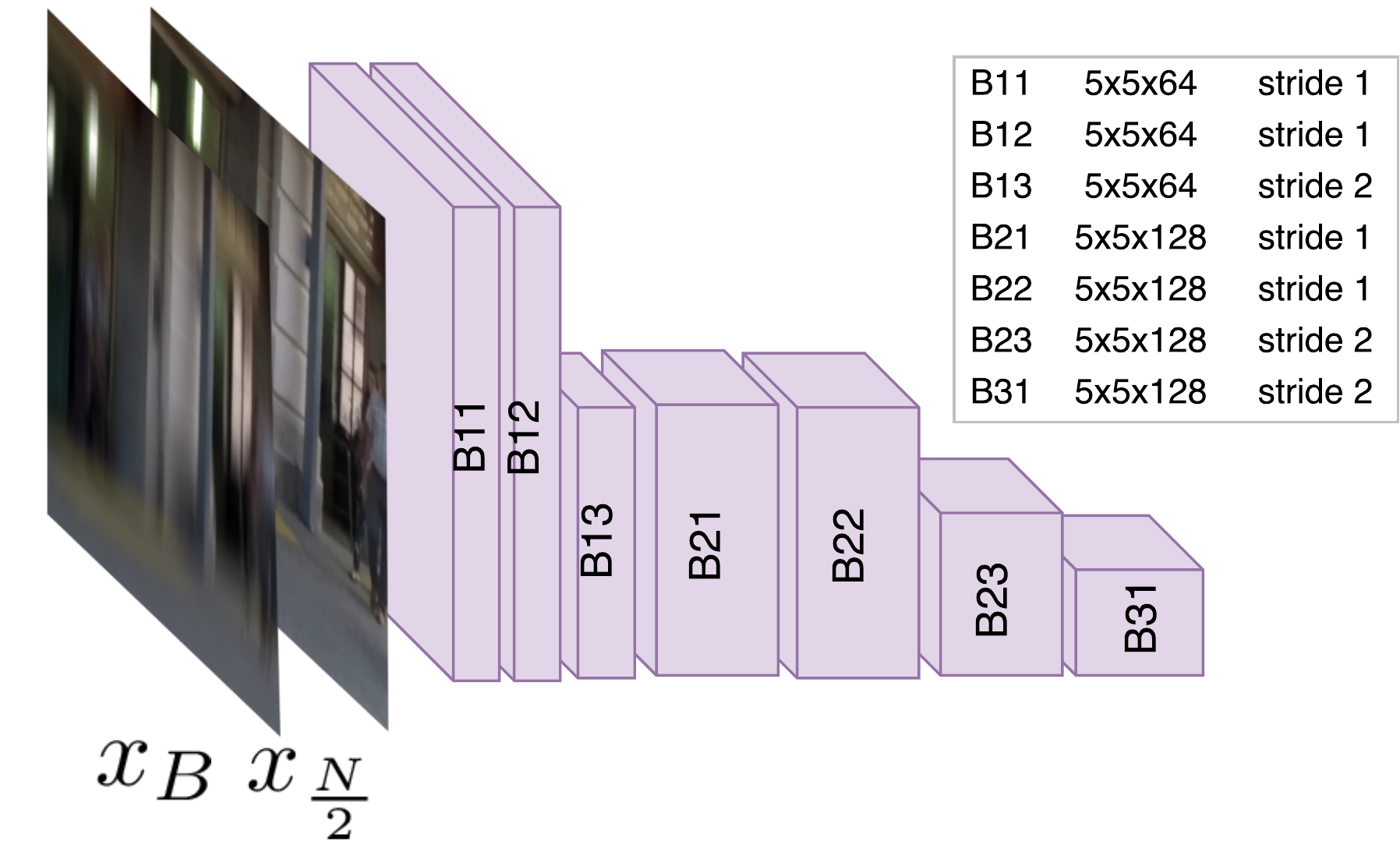}\\

            (a) RVE architecture. & (b)  BIE architecture.
       \end{tabular}
\end{center}
   \caption{Architectures of BIE and RVE. The RVE is trained to extract a motion representation from a sequence of frames while the BIE is trained to extract a motion representation from a blurred image and a sharp image.}
\label{fig:BIEnRVE}
\end{figure}

At each time-step, a frame is fed to a convolutional encoder, which generates a feature-map to be fed as input to the ConvLSTM cell. Interpreting ConvLSTM's hidden-states as a representation of motion, the kernel-size of a ConvLSTM is correlated with the speed of the motion which it can capture. Since we need to extract motion taking place within a single exposure at fine resolution, we choose a kernel-size of $3\times3$. As can be seen in Fig. \ref{fig:BIEnRVE}(a), the encoder block is made of $4$ convolutional blocks with $3\times3$ filters. The first block is a conv layer with stride of $1$ and the rest contain a conv layer with stride of $2$, followed by a Resblock. The number of feature maps in the outputs of these blocks are $16$, $32$, $64$ and $128$, respectively. A ConvLSTM cell operates on the features returned by the last block and augments it with memory from previous time-steps.

Overall, each module can be represented as $h^{enc}_n = enc(h^{enc}_{n-1},x_{n})$,
where $h^{enc}_{n}$ is encoder ConvLSTM state at time step $n$ and $x_{n}$ is the $n^{th}$ sharp frame of the video.

\subsection{Recurrent Video Decoder (RVD)}
\label{sec:rvd}

\begin{figure}
\begin{center}
   \includegraphics[width=0.45\textwidth]{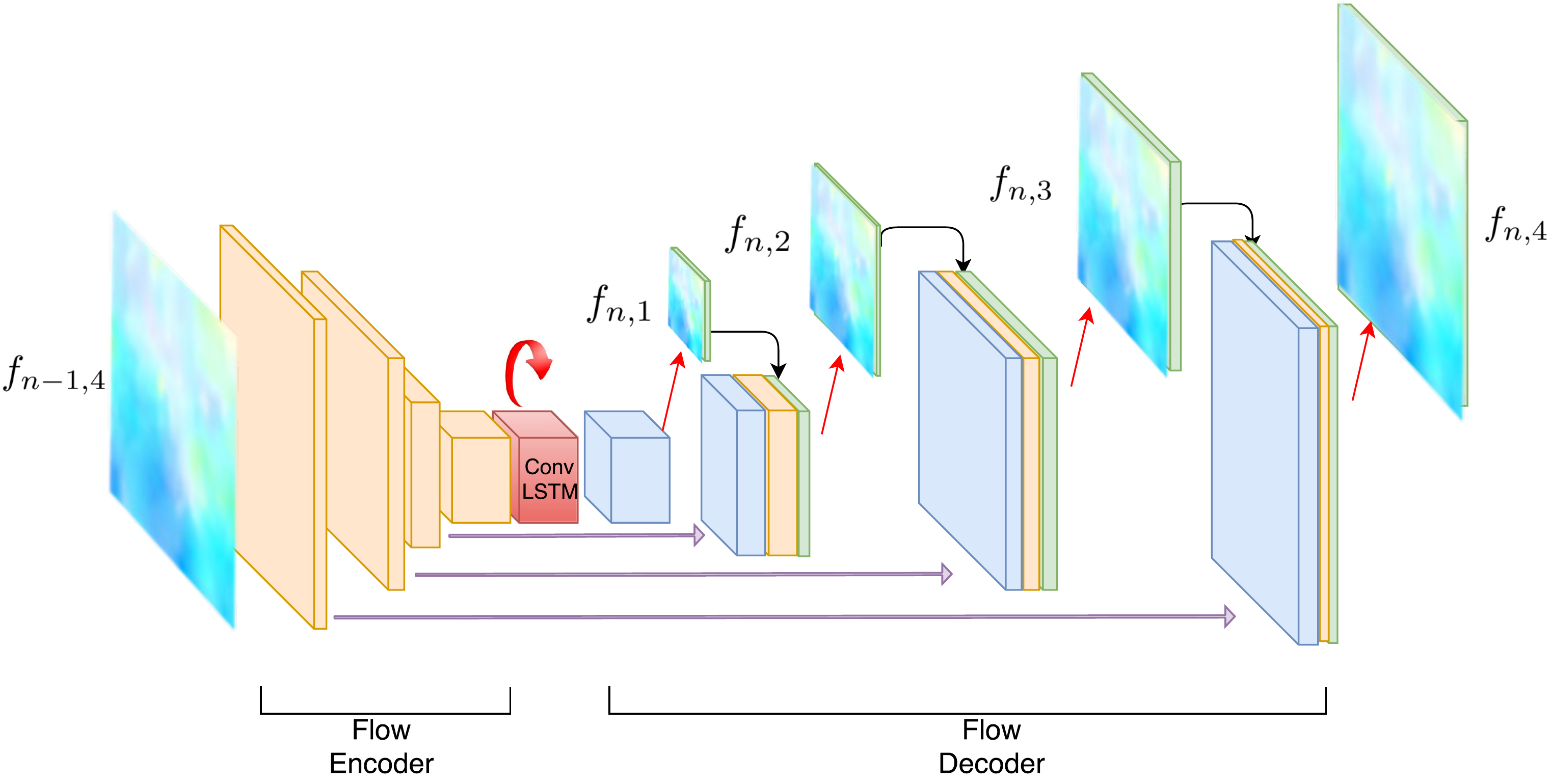}
\end{center}
\vspace{-4mm}
   \caption{Our Recurrent Video Decoder (RVD). This module recurrently generates optical flows which are warped to transform the sharp frame. Flows are estimated at 4 different scales.}
\label{fig:RVD}
\end{figure}

The task of RVD is to construct a sequence of frames using the motion representation provided by RVE and the (known) central frame ($x_{\lfloor\frac{N}{2}\rfloor}$) of the sequence. The RVD contains a flow encoder which utilizes a structure similar to the RVE. Instead of accepting images, it accepts optical flows. The flow encoding is fed to a ConvLSTM cell whose first hidden state is initialized with the last hidden state $h_{e,N}$ of the RVE. To estimate optical flows for a time-step, the output of the ConvLSTM cell is passed to a Flow decoder network ($F_D$). The flow estimated by $F_D$ at each time-step is fed to a transformer module ($T$) which returns the estimated frame $\hat{x}_{n}$. The descriptions of $F_D$ and $T$ are provided below.
\\
\textbf{Flow Decoder ($F_D$)}:
Realizing that the flow at current step is related to the previous one, 
we perform recurrence on optical flows for consecutive frames. The design of $F_D$ is illustrated in Fig. \ref{fig:RVD}. 
$F_D$ accepts the output of ConvLSTM unit at any time-step and generates a flow-map. For robust estimation, we further perform estimation of flow at multiple scales using deconvolution (deconv) layers which ``unpool'' the feature maps and increase the spatial dimensions by a factor of $2$. Inspired by \cite{ronneberger2015u}, we make use of skip connections between the layers of flow encoder and $F_D$.
All deconv operations use $ 4\times 4$ filters and the convolutional operations use $3\times3$ filters. The output of the ConvLSTM cell is passed through a convolutional layer to estimate the flow $f_{n,1}$. The cell output is also passed through a deconv layer before being concatenated with the upsampled $f_{n,1}$ and the corresponding feature-map coming from the encoder, to obtain a hybrid feature map at that scale. As shown in Fig. \ref{fig:RVD}, this process is repeated $3$ more times to obtain the flow maps at subsequently higher scales ($f_{n,2...4}$).
\\

\subsection{Blurred Image Encoder (BIE)}
\label{sec:bie}

We make use of the trained encoder-decoder couplet to solve the task of extracting video from a blurred image. 
We advocate a novel strategy of utilizing spatio-temporal embeddings to guide the training of a CNN. The trained decoder has learnt to generate optical flow for all time-steps from the encoder's hidden state. We make use of this proxy network to solve the task of blurred image to video generation.

The use of optical flow recurrence enables our network to prefer temporally consistent sequences, which preempts it from returning arbitrarily ordered frames. However, directional ambiguity stays.
For a scene with multiple objects, the ambiguity becomes more pronounced as each object can have its own independent motion. The BIE is connected with the pre-trained RVD and the pair is trained (RVD is fine-tuned) using a combination of ordering-invariant frame reconstruction loss and spatial motion smoothness loss over the RVD outputs (described later). No such ambiguity exists in the video autoencoder since the RVD has to exactly reproduce the video which is fed to RVE.

The BIE is implemented as a CNN which specializes in extracting motion features from a blurred image (we experimentally found that feeding the central sharp frame along with the blurred image improves its performance). The BIE is tasked to extract the sequential motion in the image by capturing local motion, e.g. at the smeared edges in the image. Moreover, the generated encoding should be such that the RVD can reconstruct motion trajectories. The BIE has 7 convolutional layers with kernel sizes as shown in Fig. \ref{fig:BIEnRVE}(b). Each layer (except the last) is followed by batch-normalization and leaky ReLU non-linearity.

\begin{figure}
\begin{center}
   \includegraphics[width=0.5\textwidth]{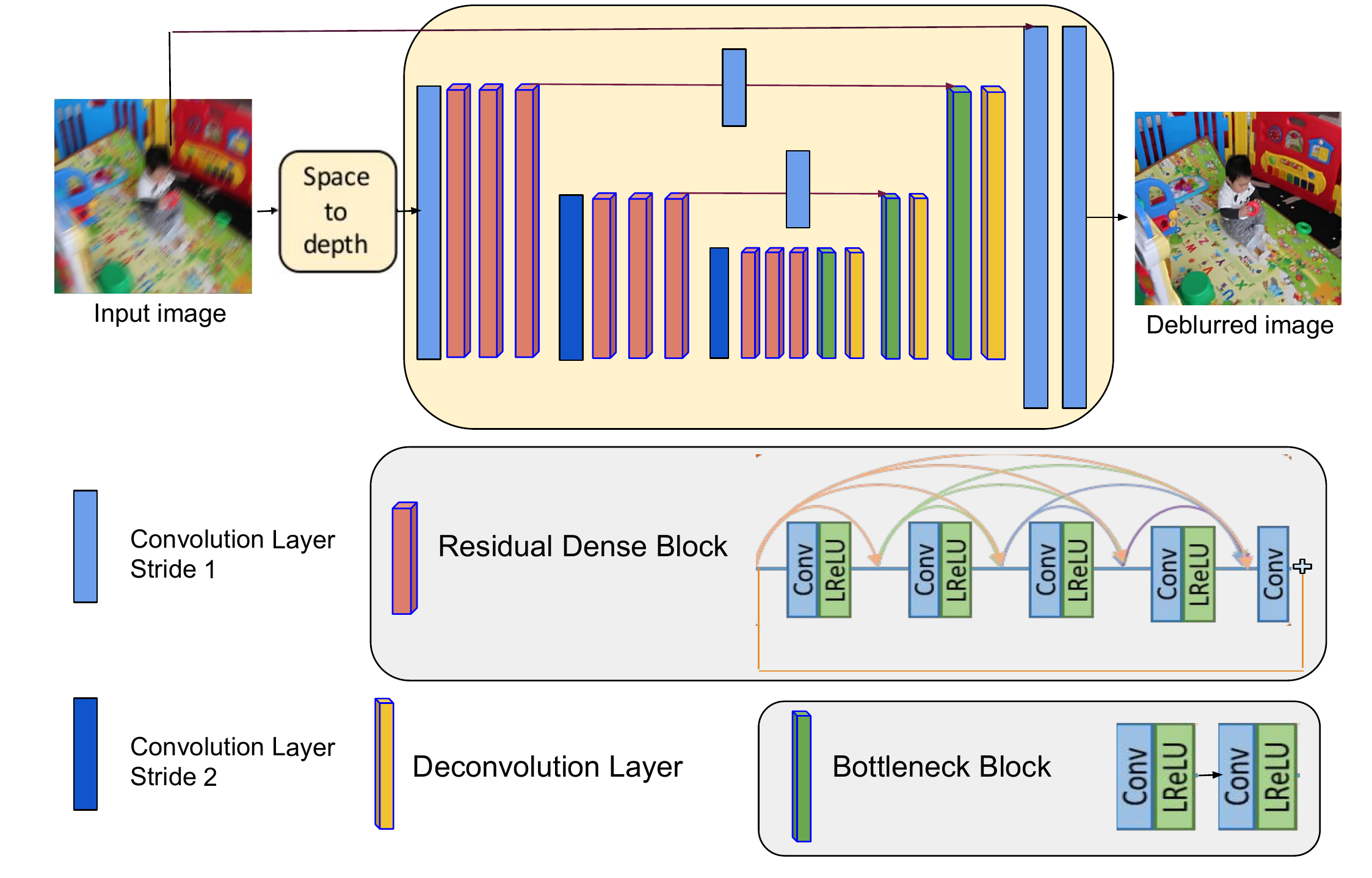}
\end{center}
\vspace{-4mm}
   \caption{An overview of our dense deblurring architecture which we utilize to estimate the central sharp frame. It follows an encoder-decoder design with residual-dense blocks, bottleneck blocks, and skip connections present at 3 different sub-scales.}
\label{fig:SID}
\end{figure}

\subsection{Deblurring Module (DM)}

We propose an independent network for deblurring the motion blurred observation. The estimated sharp frame is fed to both BIE and RVD during testing. 

  Recent works on image restoration have proposed end-to-end trainable networks which require labeled pairs of degraded and sharp images. Among them, \cite{nah2017deep,tao2018scale} have achieved promising results using multi-scale CNN composed of residual connections. We explore a more effective network architecture which is inspired by prior methods that use multi-level and multi-scale features. Our high-level design is similar to that of U-Net~\cite{ronneberger2015u}, which has been used extensively for preserving global context information in various image-to-image tasks~\cite{isola2017image}. Based on the observation that increase in number of layers and connections across them leads to a boost in feature extraction capability, the encoder structure of our network utilizes a cascade of Residual Dense Blocks (RDB)~\cite{zhang2018residual} instead of convolutional layers. An RDB is a cascade of convolutional layers connected through a rich set of residual and concatenation connections which immensely improves feature extraction capability by reusing features across multiple layers.  Inclusion of such connections maximizes information flow along the intermediate layers and results in better convergence. These units efficiently learn deeper and more complex features than a network with residual connections (which have been used extensively in recent deblurring methods\cite{nah2017deep,kupyn2017deblurgan,tao2018scale,jin2018learning}), while requiring fewer parameters.

Our proposed deblurring architecture is depicted in Fig. \ref{fig:SID}. The decoder part of our network contains $3$ pairs of up-sampling blocks to gradually enlarge the spatial resolution of feature maps. Each up-sampling block contains a bottleneck layer~\cite{jegou2017one} followed by a deconvolution layer. Each convolution layer (except the last) is followed by a non-linearity. Similar to U-Net, features corresponding to the same dimension in encoder and decoder are merged with the help of projection layers. The output of the final up-sampling block is passed through two additional convolutional layers to reconstruct the output sharp image. Our network uses an asymmetric encoder-decoder architecture, where the network capacity becomes higher benefiting from the dense connections. 

\section{Experiments}

In this section, we carry out quantitative and qualitative comparisons of our approach with state-of-the-art methods for deblurring as well as video extraction tasks.

\setlength{\tabcolsep}{1.4pt}
\begin{table}[]
\centering
\begin{tabular}{|c|c|c|c|c|c|c|c|c|}
\hline

Method & \cite{xu2013unnatural} & \cite{whyte2012non} & \cite{sun2015learning} & \cite{gong2017motion} & \cite{nah2017deep} & \cite{kupyn2017deblurgan} & \cite{tao2018scale} & Ours \\
\hline
\small{PSNR(dB)} & 21 & 24.6 & 24.5 &  26.4  & 28.9 & 27.2 & 30.10 & \textbf{30.58} \\
\small{SSIM} & 0.740 & 0.845 & 0.851 & 0.863 & 0.911 & 0.905 & 0.933 & \textbf{0.941}\\
\small{Time (s)} & 3800 & 700 & 1500 & 1200 & 6 & 0.8 & 0.4 & \textbf{0.02} \\
\small{Size(MB)} & - & - & 54.1 & 41.2 & 300 & 45.6 & 27.5 & \textbf{17.9} \\
\hline
\end{tabular}
\vspace{0mm}
\caption{Performance comparison of our deblurring network with existing methods on the benchmark dataset \cite{nah2017deep}.
\vspace{-4mm}
\label{TableGopro}}
\end{table}
\setlength{\tabcolsep}{1.4pt}

\begin{figure*}%[b!]
	\scriptsize
	\centering
	\begin{tabular}{cc}	
		\begin{adjustbox}{valign=t}
		\tiny
			\begin{tabular}{c}
				\includegraphics[width=0.22\textwidth]{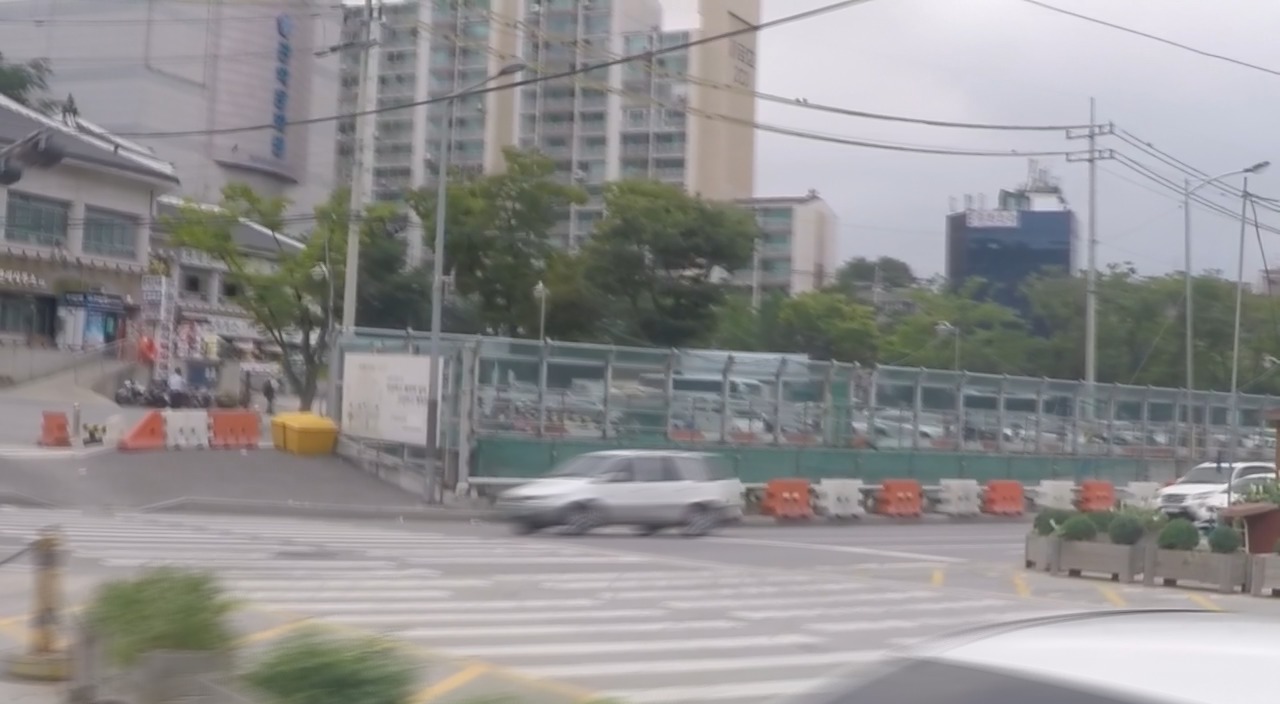}
				\\
			%	Blurred Image
				%\textsc{B100}: img_092
				
			\end{tabular}
		\end{adjustbox}
		\hspace{-1.3mm}
		\begin{adjustbox}{valign=t}
		\tiny
			\begin{tabular}{cccccc}
				\includegraphics[bb=490 170 740 280,clip=True,width=\widthscalefive \textwidth]{deblurring/000105_blurred.jpg} & 
				\includegraphics[bb=499 178 749 288,clip=True,width=\widthscalefive \textwidth]{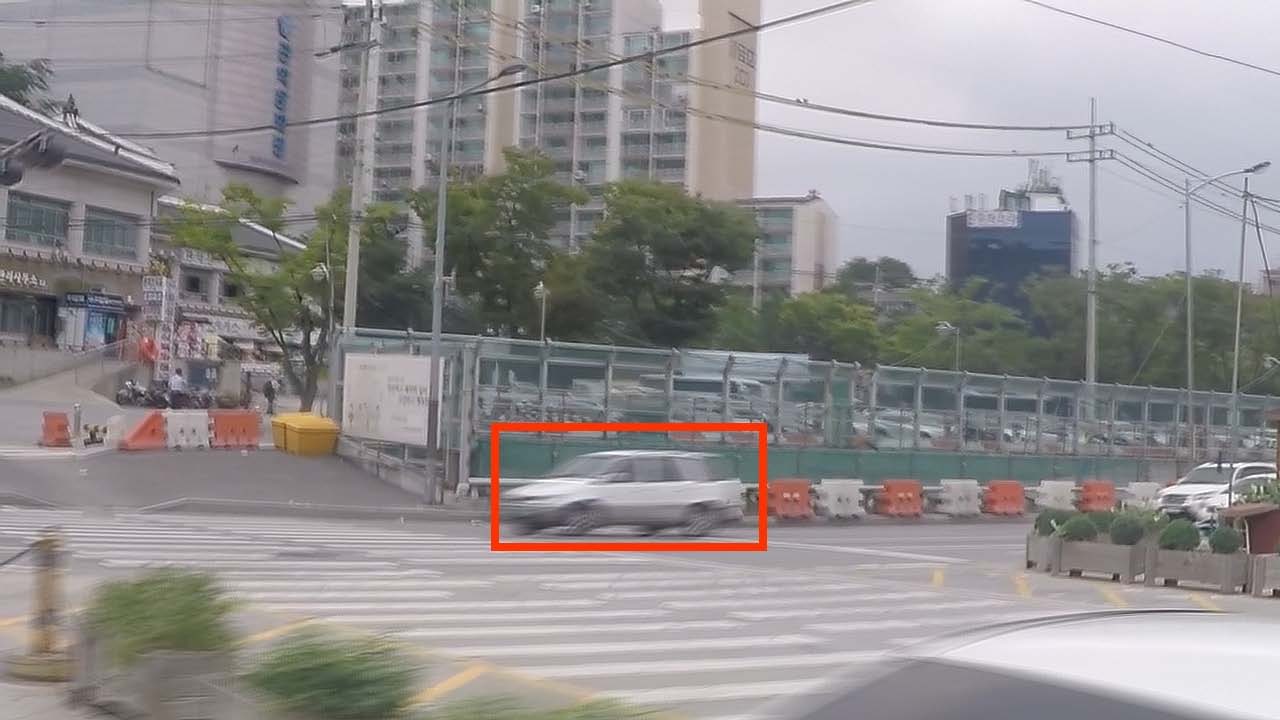} & %\hspace{\fsdttwofig} &
				\includegraphics[bb=490 180 740 290,clip=True,width=\widthscalefive \textwidth]{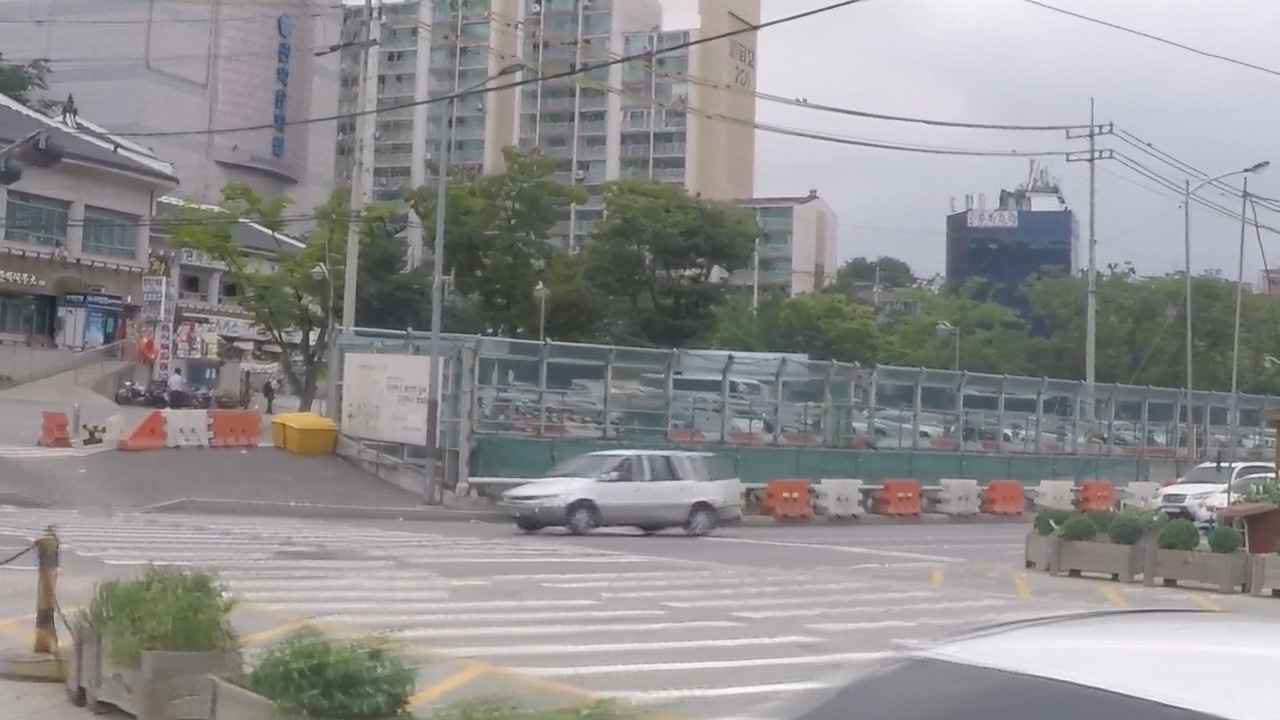} & %\hspace{\fsdttwofig} &
				\includegraphics[bb=490 170 740 280,clip=True,width=\widthscalefive \textwidth]{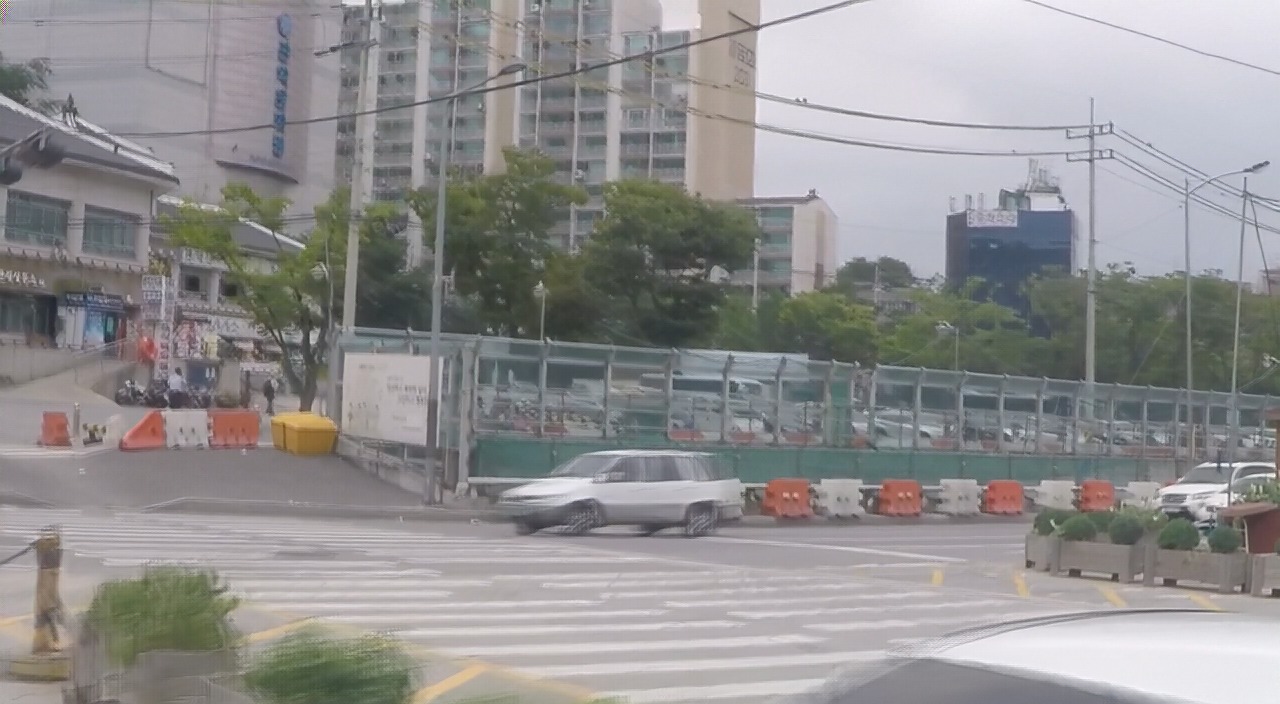} & %\hspace{\fsdttwofig} &
								\includegraphics[bb=490 170 740 280,clip=True,width=\widthscalefive \textwidth]{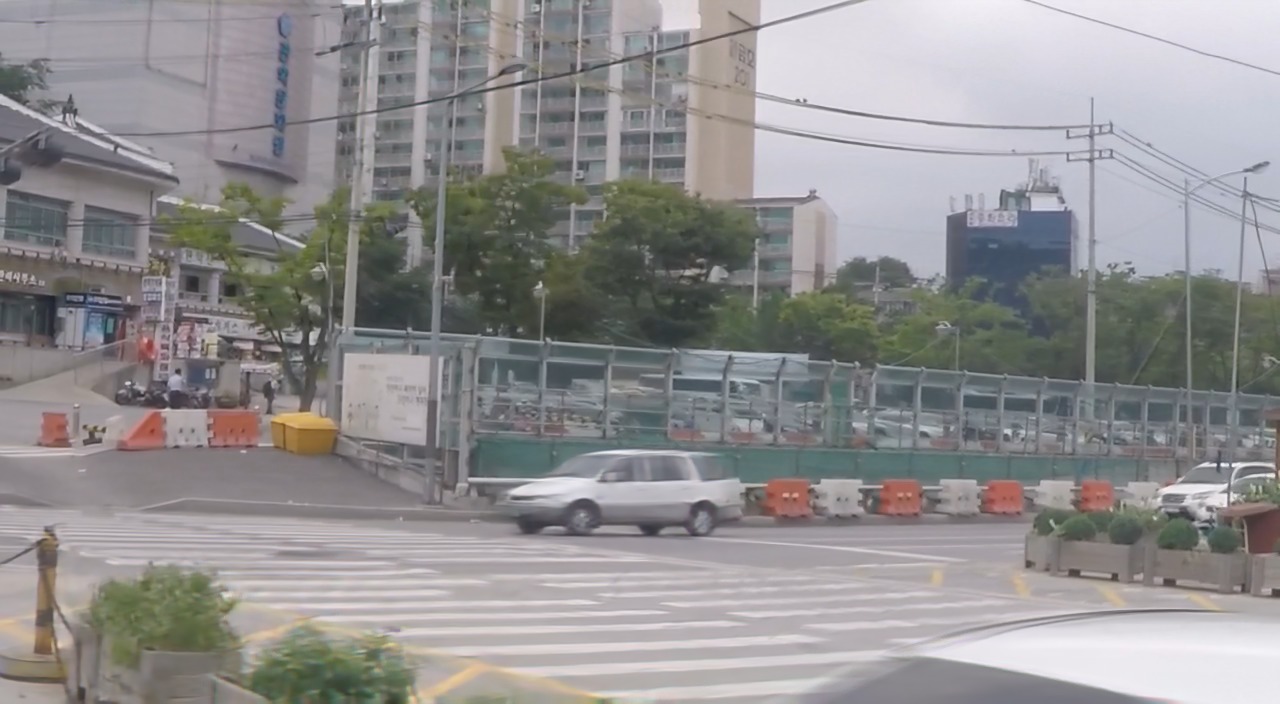} & %\hspace{\fsdttwofig} &
				\includegraphics[bb=490 170 740 280,clip=True,width=\widthscalefive \textwidth]{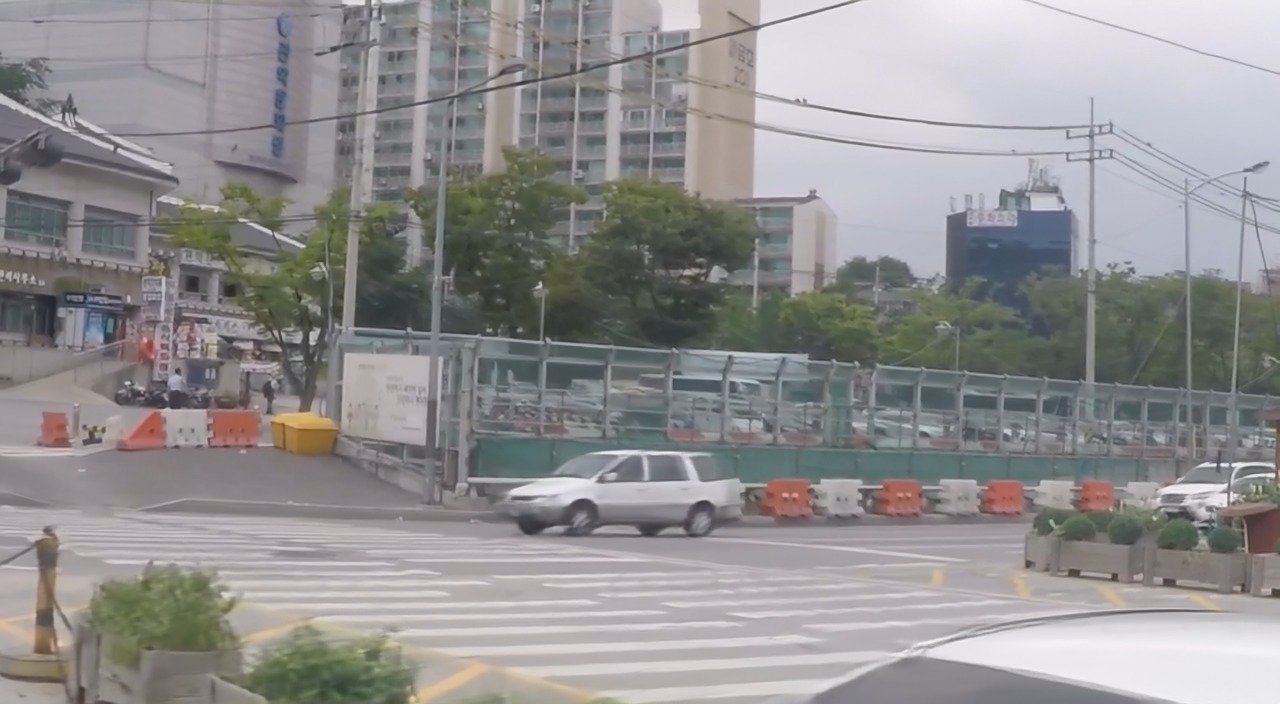}
				\\
				\includegraphics[bb=680 1 1040 150,clip=True,width=\widthscalefive \textwidth]{deblurring/000105_blurred.jpg} & %\hspace{\fsdttwofig} &
				\includegraphics[bb=680 1 1040 150,clip=True,width=\widthscalefive \textwidth]{deblurring/000105_whyte.jpg} & %\hspace{\fsdttwofig} &
				\includegraphics[bb=680 1 1040 150,clip=True,width=\widthscalefive \textwidth]{deblurring/000105_nah.jpg} & %\hspace{\fsdttwofig} &
				\includegraphics[bb=680 1 1040 150,clip=True,width=\widthscalefive \textwidth]{deblurring/000105_deblurgan.jpg} & %\hspace{\fsdttwofig} &
								\includegraphics[bb=680 1 1040 150 280,clip=True,width=\widthscalefive \textwidth]{deblurring/000105_srn.jpg} & %\hspace{\fsdttwofig} &
				\includegraphics[bb=680 1 1040 150,clip=True,width=\widthscalefive \textwidth]{deblurring/000105_ours.jpg} 
				\\
			\end{tabular}
		\end{adjustbox}
				\\			
		\begin{adjustbox}{valign=t}
		\tiny
			\begin{tabular}{c}
				\includegraphics[width=0.22\textwidth]{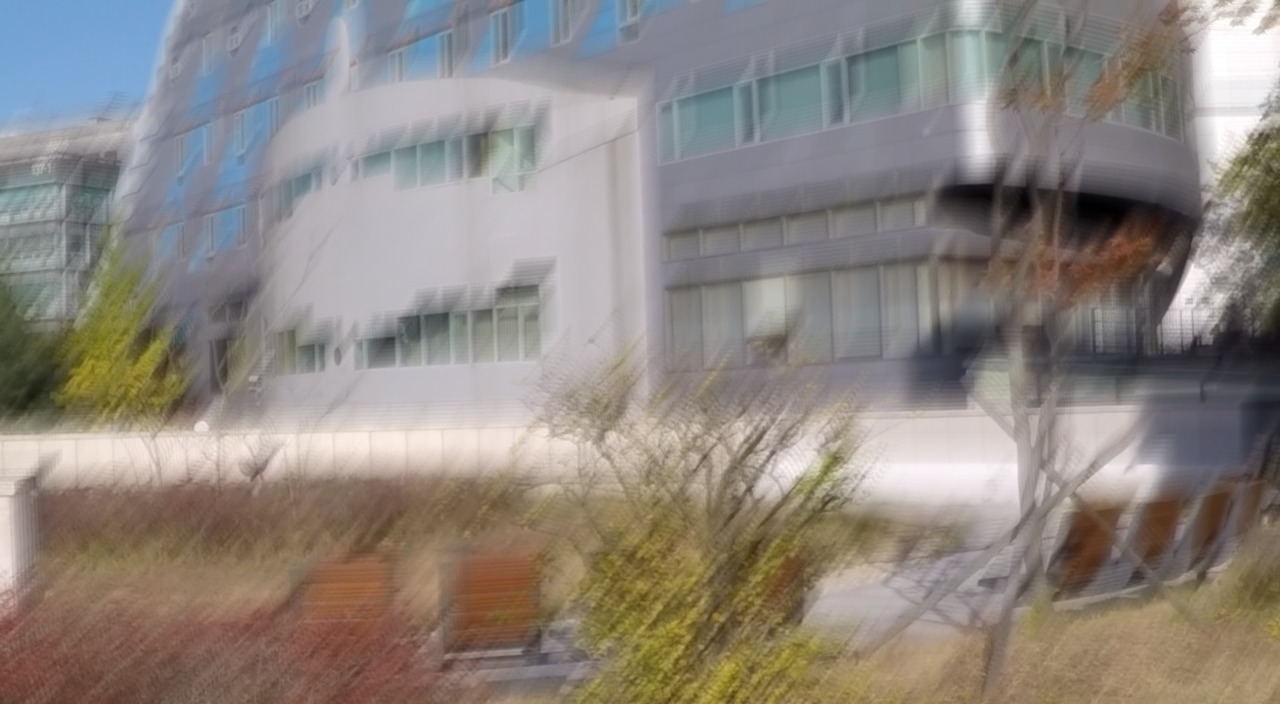}
				\\
				Blurred Image
			\end{tabular}
		\end{adjustbox}
		\hspace{-1.3mm}
		\begin{adjustbox}{valign=t}
		\tiny
			\begin{tabular}{cccccc}
				\includegraphics[bb=480 1 880 200,clip=True,width=\widthscalefive \textwidth]{deblurring/000045_blurred.jpg} & %\hspace{\fsdttwofig} &
				\includegraphics[bb=480 1 880 200,clip=True,width=\widthscalefive \textwidth]{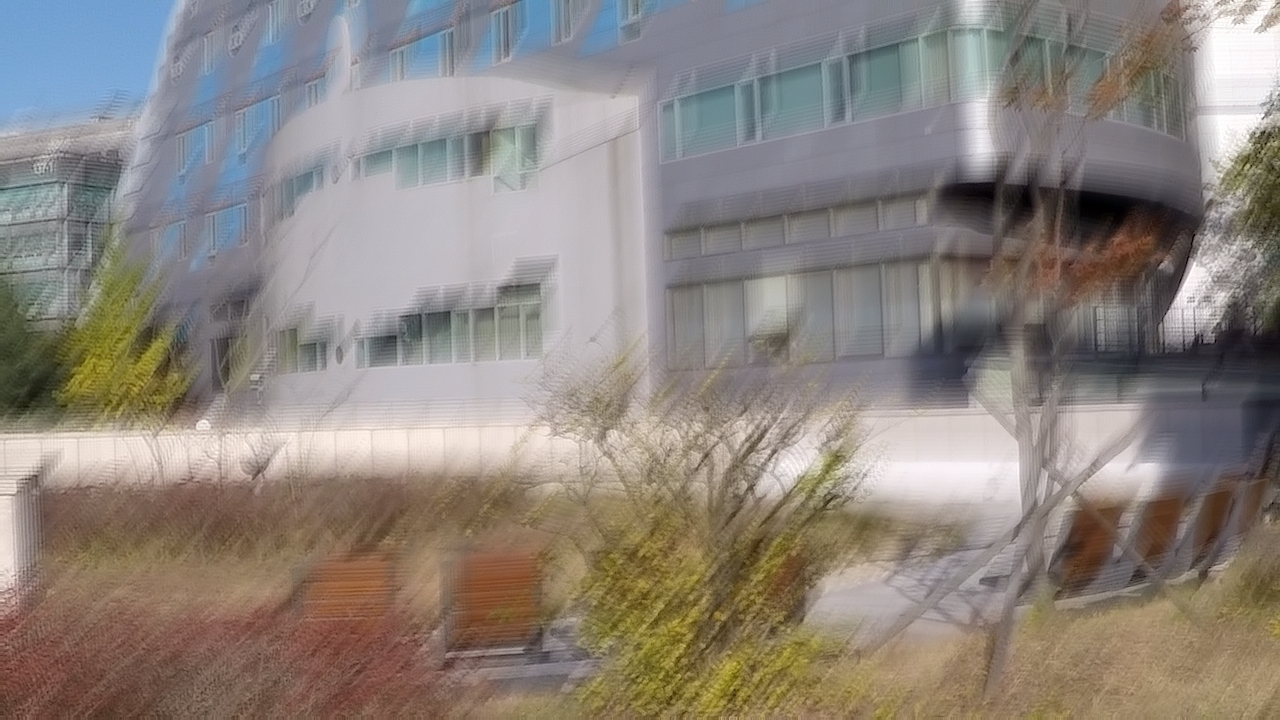} & %\hspace{\fsdttwofig} &
				\includegraphics[bb=480 1 880 200,clip=True,width=\widthscalefive \textwidth]{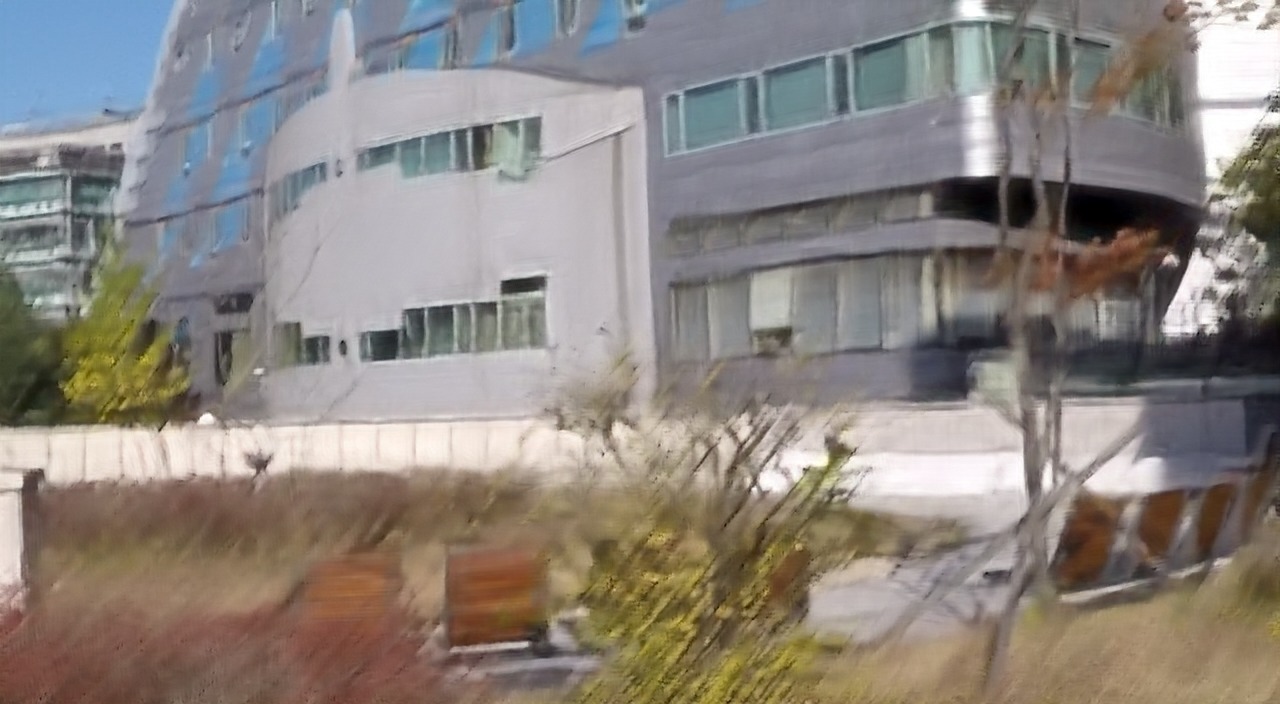} & %\hspace{\fsdttwofig} &
				\includegraphics[bb=480 1 880 200,clip=True,width=\widthscalefive \textwidth]{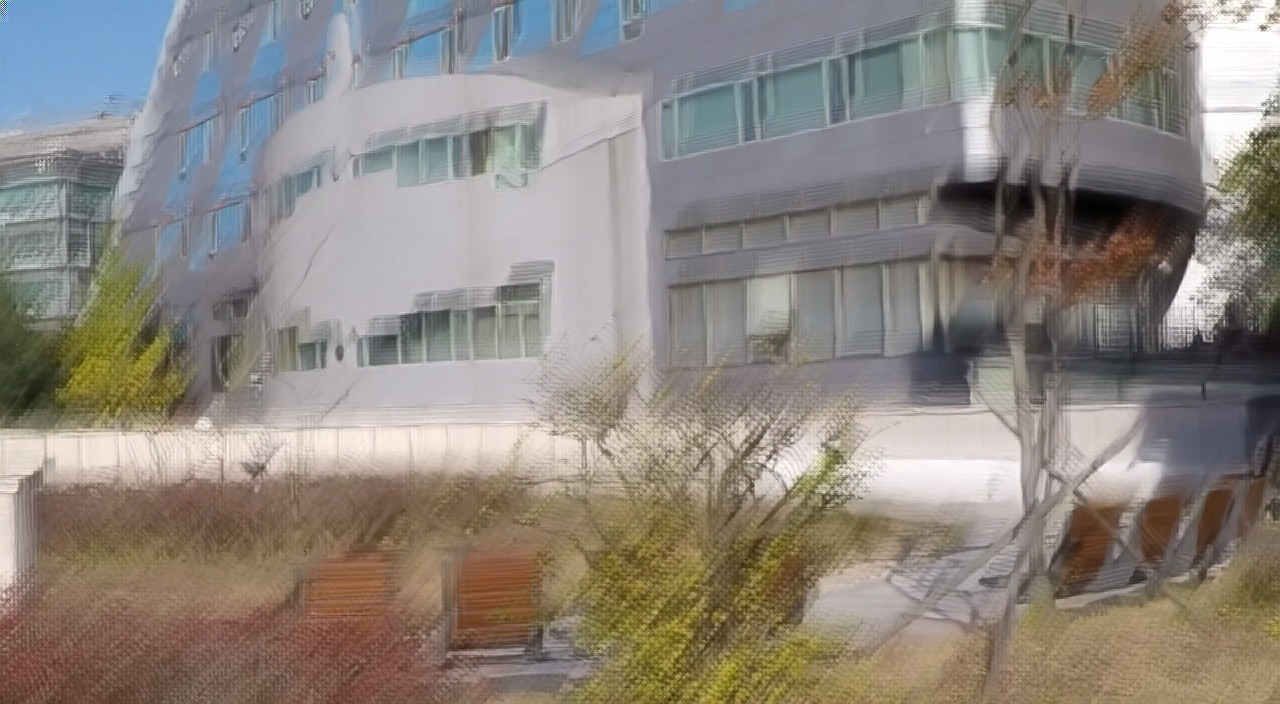} & %\hspace{\fsdttwofig} &
								\includegraphics[bb=480 1 880 200 280,clip=True,width=\widthscalefive \textwidth]{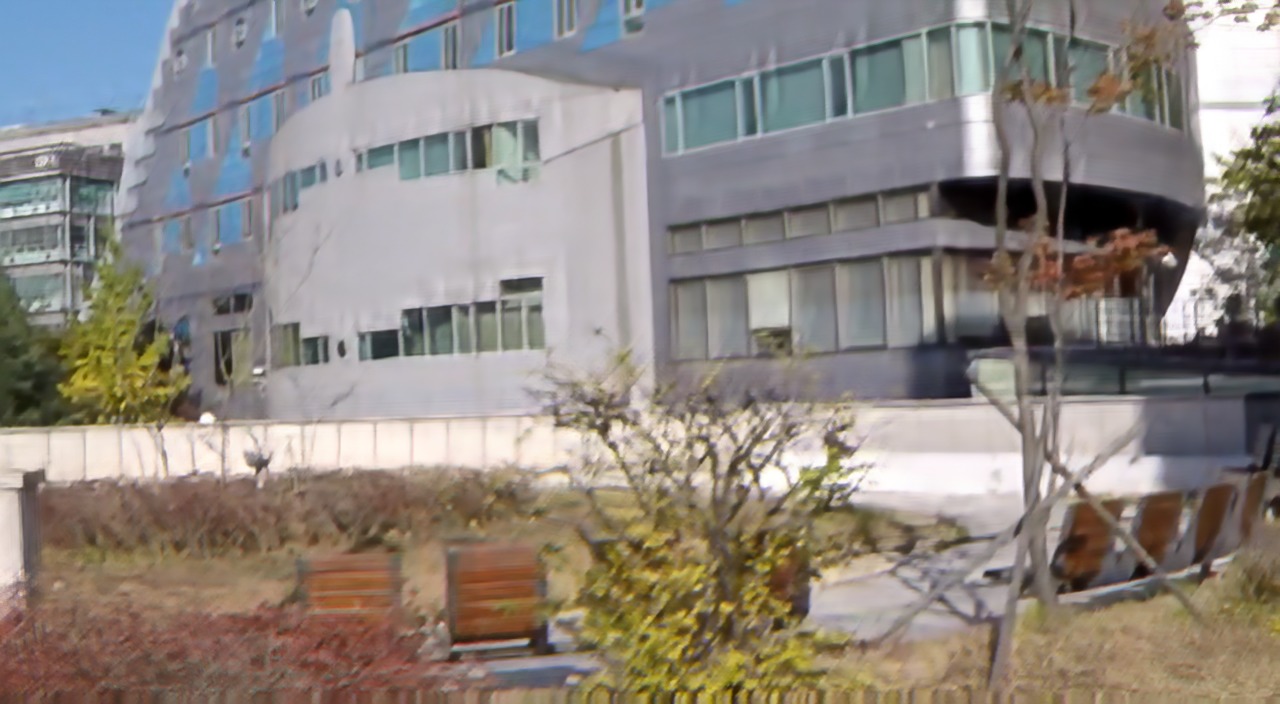} & %\hspace{\fsdttwofig} &3
				\includegraphics[bb=480 1 880 200,clip=True,width=\widthscalefive \textwidth]{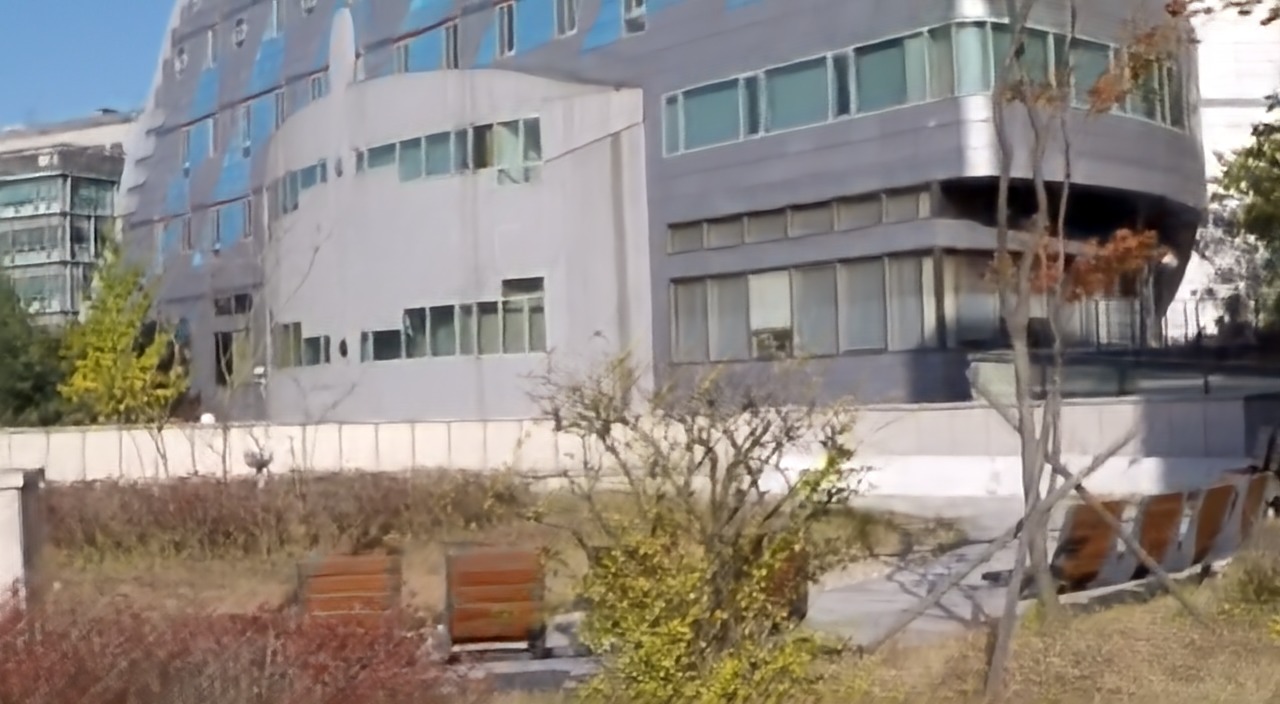} 				\vspace{-3mm}

				\\ 
				\includegraphics[bb=380 400 980 780,clip=True,width=\widthscalefive \textwidth]{deblurring/000045_blurred.jpg} & %\hspace{\fsdttwofig} &
				\includegraphics[bb=380 400 980 780,clip=True,width=\widthscalefive \textwidth]{deblurring/000045_whyte.jpg} & %\hspace{\fsdttwofig} &
				\includegraphics[bb=380 400 980 780,clip=True,width=\widthscalefive \textwidth]{deblurring/000045_nah.jpg} & %\hspace{\fsdttwofig} &
				\includegraphics[bb=380 400 980 780,clip=True,width=\widthscalefive \textwidth]{deblurring/000045_deblurgan.jpg} & %\hspace{\fsdttwofig} &
								\includegraphics[bb=380 400 980 780 280,clip=True,width=\widthscalefive \textwidth]{deblurring/000045_srn.jpg} & %\hspace{\fsdttwofig} &
				\includegraphics[bb=380 400 980 780,clip=True,width=\widthscalefive \textwidth]{deblurring/000045_ours.jpg} 
				\\ 
				Blurred patch& %\hspace{\fsdttwofig} &
				Whyte \etal~\cite{whyte2012non} & %\hspace{\fsdttwofig} &
				Nah \etal~\cite{nah2017deep} & %\hspace{\fsdttwofig} &
				DelurGAN~\cite{kupyn2017deblurgan}& % \hspace{\fsdttwofig} &
				SRN~\cite{tao2018scale}& %\hspace{\fsdttwofig} &
				Ours%\hspace{\fsdttwofig} &
				\\
			\end{tabular}
		\end{adjustbox}	
	\end{tabular}
	\caption{Visual comparisons of deblurring results on test dataset~\cite{nah2017deep} (best view in high resolutions).}
\label{fig:comp_deep}
\end{figure*}

 \begin{figure*}[t]
 \begin{center}
 \begin{tabular}{cccccc}
   \includegraphics[width=\widthscalefour\textwidth]{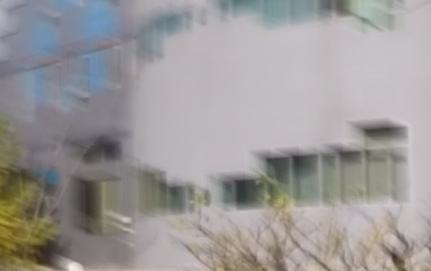} &
   \includegraphics[width=\widthscalefour\textwidth]{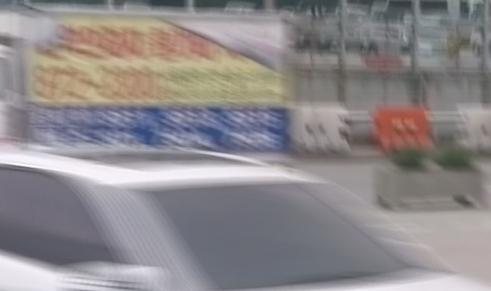} &
      \includegraphics[width=\widthscalefour\textwidth]{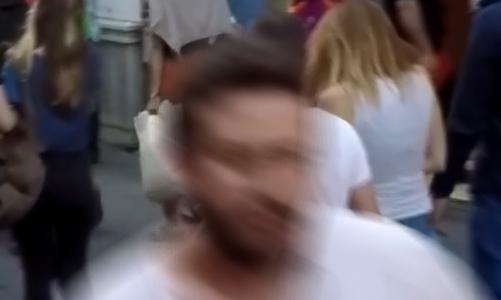} &
\includegraphics[width=\widthscalefour\textwidth]{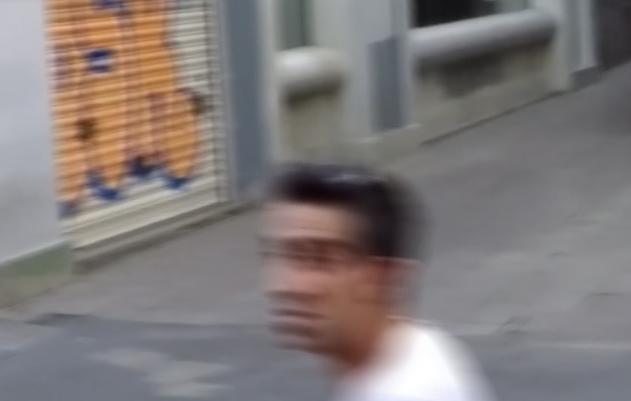}  &
   \includegraphics[width=\widthscalefour\textwidth]{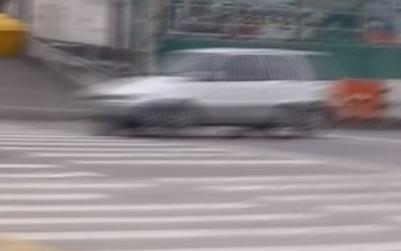} &       
          \includegraphics[width=\widthscalefour\textwidth]{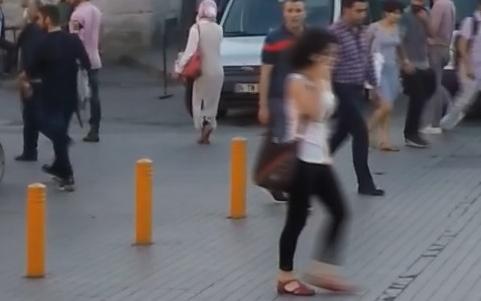}
           \\

   \animategraphics[width=\widthscalefour\textwidth   ,loop]{9}{"NEWcropped/ours1/"}{0}{8}&

   \animategraphics[width=\widthscalefour\textwidth    ,loop]{9}{"NEWcropped/ours3/"}{0}{8} &
      \animategraphics[width=\widthscalefour\textwidth    ,loop]{9}{"NEWcropped/ours4/"}{0}{8} &  
        \animategraphics[width=\widthscalefour\textwidth    ,loop]{9}{"NEWcropped/ours5/"}{0}{8}   &
   \animategraphics[width=\widthscalefour\textwidth    ,loop]{9}{"NEWcropped/ours2/"}{0}{8} &        
           \animategraphics[width=\widthscalefour\textwidth    ,loop]{9}{"NEWcropped/ours6/"}{0}{8}  \\

   \animategraphics[width=\widthscalefour\textwidth   ,loop]{8}{"NEWcropped/favaro1/000066-esti"}{1}{7}&
   \animategraphics[width=\widthscalefour\textwidth    ,loop]{8}{"NEWcropped/favaro3/000229-esti"}{1}{7} & 
      \animategraphics[width=\widthscalefour\textwidth    ,loop]{8}{"NEWcropped/favaro4/004039-esti"}{1}{7} & 
         \animategraphics[width=\widthscalefour\textwidth    ,loop]{8}{"NEWcropped/favaro5/003069-esti"}{1}{7} &  
   \animategraphics[width=\widthscalefour\textwidth    ,loop]{8}{"NEWcropped/favaro2/000108-esti"}{1}{7} &         
            \animategraphics[width=\widthscalefour\textwidth    ,loop]{8}{"NEWcropped/favaro6/000040-esti"}{1}{7}   \\         
   (a) & (b) & (c) & (d) & (e) & (f)
\end{tabular}
   \caption{Comparisons of our video extraction results with \cite{jin2018learning} on motion blurred images obtained from the test dataset of \cite{nah2017deep}. The first row shows the blurred images while the second and third rows show videos generated by our method and \cite{jin2018learning}, respectively. \emph{Videos can be viewed by clicking
on the image, when document is opened in Adobe Reader}.}  
\label{fig:gopropp}
\vspace{-5mm}
   \end{center}
\end{figure*}

\begin{figure*}[]
\begin{center}
\begin{tabular}{cccccc}
   \includegraphics[width=0.14\textwidth]{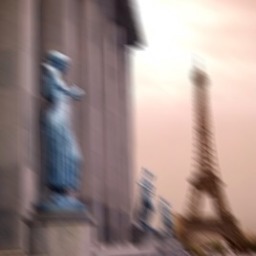} &
      \includegraphics[width=0.14\textwidth]{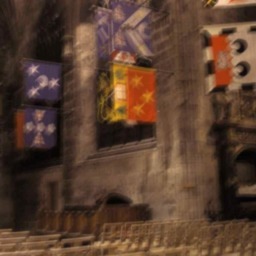} &
         \includegraphics[width=0.14\textwidth]{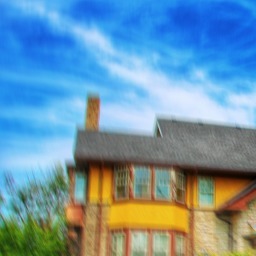} &
   \includegraphics[width=0.14\textwidth]{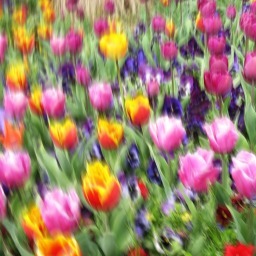} &
      \includegraphics[width=0.14\textwidth]{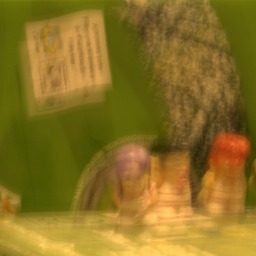}&
         \includegraphics[width=0.14\textwidth]{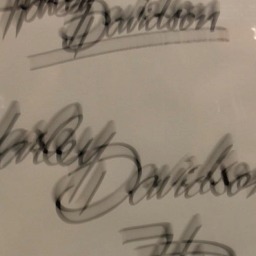}\\% &\linebreak%\vspace{1mm}
         \animategraphics[width=0.14\textwidth,loop]{8}{"3d/video/"}{2}{8}&
         \animategraphics[width=0.14\textwidth,loop]{8}{"kohler/video/"}{2}{8} &
         \animategraphics[width=0.14\textwidth,loop]{8}{"KaiSynth/house/video/"}{2}{8} &
            \animategraphics[width=0.14\textwidth,loop]{8}{"KaiSynth/flower/video/"}{2}{8} &
            \animategraphics[width=0.14\textwidth    ,loop]{8}{"KaiDataset/1/video/"}{2}{8}    &    
      \animategraphics[width=0.14\textwidth,loop]{8}{"KaiDataset/3/video/"}{2}{8}\\
   (a) & (b) & (c) & (d) & (e) & (f)
\end{tabular}  
\vspace{-1mm}
   \caption{Video generation from images blurred with global camera motion from datasets of \cite{gong2017motion},\cite{kohler2012recording} and \cite{lai2016comparative}. First row shows the blurred images. The generated videos using our method are shown in second row.}
\label{fig:cameramotion}
\end{center}
\vspace{-1.5em}
\end{figure*}

 \begin{figure*}[t]
 \begin{center}
 \begin{tabular}{ccccccc}
   \includegraphics[width=0.13\textwidth]{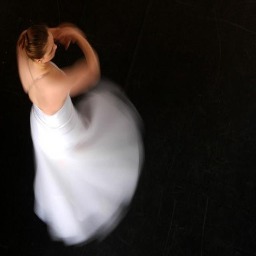} &
   \includegraphics[width=0.13\textwidth]{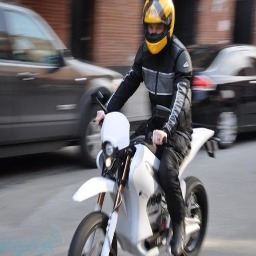} 8&

   \includegraphics[width=0.13\textwidth]{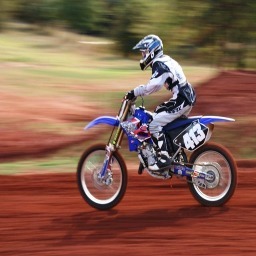} &

   \includegraphics[width=0.13\textwidth]{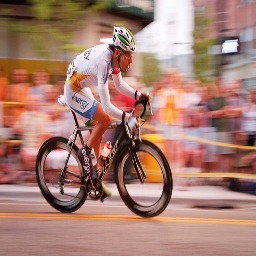} &
   \includegraphics[width=0.13\textwidth]{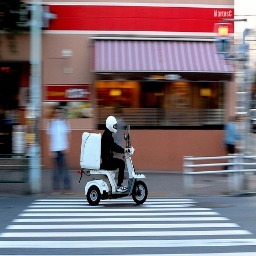} &   
      \includegraphics[width=0.13\textwidth]{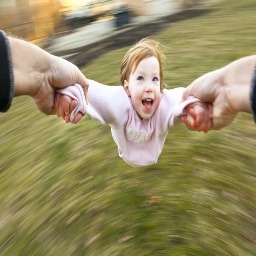} &
   \includegraphics[width=0.13\textwidth]{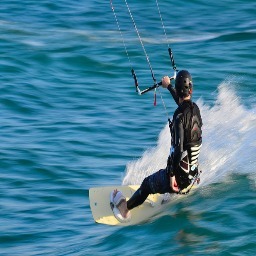}
           \\
\vspace{0mm}
   \animategraphics[width=0.13\textwidth    ,loop]{8}{"BlurDetectionDataset/im8/video/"}{2}{8}&
   \animategraphics[width=0.13\textwidth    ,loop]{8}{"BlurDetectionDataset/im12/video/"}{2}{8} &
   \animategraphics[width=0.13\textwidth    ,loop]{8}{"BlurDetectionDataset/im45/video/"}{2}{8} &
   \animategraphics[width=0.13\textwidth    ,loop]{8}{"BlurDetectionDataset/im256/video/"}{2}{8}&
   \animategraphics[width=0.13\textwidth   ,loop]{8}{"BlurDetectionDataset/im232/video/"}{2}{8} &
      \animategraphics[width=0.13\textwidth   ,loop]{8}{"BlurDetectionDataset/im272/video/"}{2}{8} &
   \animategraphics[width=0.13\textwidth   ,loop]{8}{"BlurDetectionDataset/im103/video/"}{2}{8}\\

\end{tabular}
   \caption{Video generation results on real motion blurred images from dataset of \cite{shi2014discriminative}. The first row shows the blurred images. Second row contains the extracted videos with our method.}
   \label{fig:blurdetection}
\vspace{-6mm}
   \end{center}

\end{figure*}

\subsection{Results and Comparisons for Video Extraction}
\label{favarocomparisons}

In Fig  \ref{fig:gopropp}, we give results on standard test blurred images from the dataset of \cite{nah2017deep}. Note that some of them suffer from significant blur. Fig. \ref{fig:gopropp}(a) shows an image of a planar scene which is blurred due to dominant camera motion. Fig. \ref{fig:gopropp}(b) shows a 3D scene blurred due to camera motion. Figs. \ref{fig:gopropp}(c-f) show results on blurred images with dynamic object motion. Observe that the videos generated by our approach are realistic and qualitatively consistent with the blur and depth of the scene, even when the foreground incurs large motion. Our network is able to reconstruct videos from blurred images with diverse motion and scene content.

In comparison, the results of \cite{jin2018learning} suffer from local errors in deblurring, inconsistent motion estimation, as well as color distortions. We have observed that in general the method of \cite{jin2018learning} fails in cases involving high blur as direct image regression becomes difficult for large motion. In contrast, we divide the overall problem into two sub-tasks of deblurring and motion extraction. This simplifies learning and yields improvement in deblurring quality as well as motion estimation. The color issue in \cite{jin2018learning} can be attributed to the design of their networks, wherein feature extraction and reconstruction branches are different for different color channels. Our method applies the same motion to each color channel. By having a single recurrent network to generate the video, our network can be directly trained to extract even higher number of frames ($>9$) without any design change or additional parameters. In contrast, \cite{jin2018learning} requires training of an additional network for each new pair of frames. Our overall architecture is more compact ($34$ MB vs $70$ MB) and much faster ($0.02$s vs $0.45$s for deblurring and $0.39$s vs $1.10$s for video generation) as compared to \cite{jin2018learning}. 

To perform quantitative comparisons with \cite{jin2018learning}, we also trained another version of our network on the restricted case of blurred images produced by averaging $7$ successive sharp frames. For testing, $250$ blurred images of resolution $1280 \times 704$ were created using the $11$ test videos from the dataset of \cite{nah2017deep}. We compared the videos estimated by the two methods using the ambiguity invariant loss function. The average error was found to be $49.06$ for \cite{jin2018learning} and $44.12$ for our method. Thus, even for the restricted case of small blur, our method performs favorably. Repeating the same experiment for $9$ frames (i.e. for large blur from the same test videos) led to an error of $48.24$ for our method, which is still less than the 7-frame error of \cite{jin2018learning}. We could not compute the 9-frame error for \cite{jin2018learning} as their network is rigidly designed for $7$ frames only.
\section{Conclusions}
We introduced a new methodology for video generation from a single blurred image. We proposed a spatio-temporal video auto-encoder based on an end-to-end differentiable architecture that learns motion representation from sharp videos in a self-supervised manner. The network predicts a sequence of optical flows and employs them to transform a sharp central frame and return a smooth video. Using the trained video decoder, we trained a blurred image encoder to extract a representation from a single blurred image, that mimics the representation returned by the video encoder. This when fed to the decoder returns a plausible sharp video representing the action within the blurred image. We also proposed an efficient deblurring architecture composed of densely connected layers that yields state-of-the-art results. The potential of our work can be extended in a variety of directions including blur-based segmentation, video deblurring, video interpolation, action recognition etc. 
Refined and complete version of this work appeared in CVPR 2019.

{\small
\bibliographystyle{ieee}
\bibliography{ref}
}

\end{document}